\documentclass[11pt,letterpaper]{article}
\usepackage{naaclhlt2016}
\usepackage{times}
\usepackage{latexsym}

\naaclfinalcopy 
\def\naaclpaperid{149} 

\usepackage{url}
\usepackage{array}
\usepackage{graphicx}
\usepackage{wrapfig}
\usepackage{subfigure}
\usepackage{color,xcolor,colortbl}
\usepackage{amsmath}
\usepackage{amssymb}
\usepackage{amsfonts,eucal,amsbsy,amsthm,amsopn}
\usepackage{sidecap}
\usepackage{multirow}
\usepackage{algorithm}
\usepackage{algorithmic}

\makeatletter
\newcommand{\longdash}[1][2em]{%
  \makebox[#1]{$\m@th\smash-\mkern-7mu\cleaders\hbox{$\mkern-2mu\smash-\mkern-2mu$}\hfill\mkern-7mu\smash-$}}
\makeatother
\newcommand{\omitskip}{\kern-\arraycolsep}

\title{Automatic Summarization of Student Course Feedback}
\author{Wencan Luo$^\dagger$ \quad Fei Liu$^\ddagger$ \quad Zitao Liu$^\dagger$ \quad Diane Litman$^\dagger$ \\
  $^\dagger$University of Pittsburgh, Pittsburgh, PA 15260 \\
  $^\ddagger$University of Central Florida, Orlando, FL 32716 \\
  {\tt \{wencan, ztliu, litman\}@cs.pitt.edu} \quad {\tt feiliu@cs.ucf.edu}\\}

\date{}

\begin{document}
\maketitle

\begin{abstract}

Student course feedback is generated daily in both classrooms and online course discussion forums.
Traditionally, instructors manually analyze these responses in a costly manner.
In this work, we propose a new approach to summarizing student course feedback based on the integer linear programming (ILP) framework.
Our approach allows different student responses to share co-occurrence statistics and alleviates sparsity issues.
Experimental results on a student feedback corpus show that our approach 
outperforms a range of baselines in terms of both ROUGE scores and human evaluation.
\end{abstract}

\section{Introduction}
\label{sec:intro}

Instructors love to solicit feedback from students.
Rich information from student responses can reveal complex teaching problems, help teachers adjust their teaching strategies, and create more effective teaching and learning experiences.
Text-based student feedback is often manually analyzed by teaching evaluation centers in a costly manner.
Albeit useful, the approach does not scale well. 
It is therefore desirable to automatically summarize the student feedback produced in online and offline environments.
In this work, student responses are collected from an introductory materials science and engineering course, taught in a classroom setting.
Students are presented with prompts after each lecture and asked to provide feedback. These prompts solicit ``\textit{reflective feedback}'' \cite{boud:2013} from the students. An example is presented in Table \ref{tab:example}.

\begin{table}[t]
\setlength{\tabcolsep}{5pt}
\renewcommand{\arraystretch}{1.0}
\begin{footnotesize}
\begin{tabular}{ l }
\hline
\rule{0pt}{3ex}\textbf{Prompt}\\
Describe what you found most interesting in today's class\\
\hline
\rule{0pt}{3ex}\textbf{Student Responses}\\
S1: The main topics of this course seem interesting and \\
\quad \,\, correspond with my major (Chemical engineering)\\
S2: I found the group activity most interesting \\
S3: Process that make materials \\
S4: I found the properties of bike elements to be most \\
\quad \,\, interesting\\
S5: How materials are manufactured\\
S6: Finding out what we will learn in this class was \\
\quad \,\, interesting to me\\
S7: The activity with the bicycle parts\\
S8: ``part of a bike" activity \\
\rule[-1ex]{0pt}{0pt}... (\textit{rest omitted, 53 responses in total.})\\
\hline
\rule{0pt}{3ex}\textbf{Reference Summary}\\
- group activity of analyzing bicycle's parts\\
- materials processing\\
\rule[-1ex]{0pt}{0pt}- the main topic of this course\\

\hline
\end{tabular}
\caption{Example student responses and a reference summary created by the teaching assistant. `S1'--`S8' are student IDs.}
\label{tab:example}
\end{footnotesize}
\vspace{-0.1in}
\end{table}

In this work, we aim to summarize the student responses.
This is formulated as an extractive summarization task, where a set of representative sentences are extracted from student responses to form a textual summary. 
One of the challenges of summarizing student feedback is its lexical variety.
For example, in Table~\ref{tab:example}, ``bike elements" (S4) and ``bicycle parts" (S7), ``the main topics of this course" (S1) and ``what we will learn in this class" (S6) are different expressions that communicate the same or similar meanings.
In fact, we observe 97\% of the bigrams appear only once or twice in the student feedback corpus (\S\ref{sec:data}), 
whereas in a typical news dataset (DUC 2004), it is about 80\%. 
To tackle this challenge, we propose a new approach to summarizing student feedback, which extends the standard ILP framework by approximating the co-occurrence matrix using a low-rank alternative. 
The resulting system allows sentences authored by different students to share co-occurrence statistics.
For example, ``The activity with the bicycle parts" (S7) will be allowed to partially contain ``bike elements" (S4) although the latter did not appear in the sentence. 
Experiments show that our approach produces better results on the student feedback summarization task in terms of both ROUGE scores and human evaluation.

\section{ILP Formulation}
\label{sec:ilp}

Let $\mathcal{D}$ be a set of student responses that consist of $M$ sentences in total. Let $y_j \in \{0,1\}$, $j = \{1,\cdots, M\}$ indicate if a sentence $j$ is selected ($y_j = 1$) or not ($y_j = 0$) in the summary. 
Similarly, let $N$ be the number of unique concepts in $\mathcal{D}$. 
$z_i \in \{0,1\}$, $i = \{1, \cdots, N\}$ indicate the appearance of concepts in the summary. 
Each concept $i$ is assigned a weight of $w_i$, often measured by the number of sentences or documents that contain the concept.
The ILP-based summarization approach~\cite{Gillick:2009:NAACL} searches for an optimal assignment to the sentence and concept variables so that the selected summary sentences maximize coverage of important concepts.
The relationship between concepts and sentences is captured by a co-occurrence matrix $A \in \mathbb{R}^{N \times M}$, where $A_{ij}=1$ indicates the $i$-th concept appears in the $j$-th sentence, and $A_{ij} = 0$ otherwise. In the literature, bigrams are frequently used as a surrogate for concepts~\cite{Gillick:2008,Berg-Kirkpatrick:2011}. 
We follow the convention and use `concept' and `bigram' interchangeably in the paper.
\begin{align}
\max _{\boldsymbol{y},\boldsymbol{z}} & \textstyle \quad \sum_{i=1}^{N} w_i z_i
\label{eqn:orig_ilp_obj}\\[0.6em]
s.t. & \textstyle \quad \sum_{j=1}^{M} A_{ij} \, y_j \ge z_i 
\label{eqn:orig_ilp_bigram}\\[0.3em]
& \quad A_{ij} \, y_j \le z_i 
\label{eqn:orig_ilp_sentence}\\[0.3em]
& \textstyle \quad \sum_{j=1}^{M} l_j y_j \le L 
\label{eqn:orig_ilp_length}\\[0.3em]
& \quad y_j \in \{0,1\}, z_i \in \{0,1\} 
\label{eqn:orig_yz}
\end{align}

Two sets of linear constraints are specified to ensure the ILP validity:
(1) a concept is selected if and only if at least one sentence carrying it has been selected (Eq.~\ref{eqn:orig_ilp_bigram}), 
and (2) all concepts in a sentence will be selected if that sentence is selected (Eq.~\ref{eqn:orig_ilp_sentence}).
Finally, the selected summary sentences are allowed to contain a total of $L$ words or less (Eq.~\ref{eqn:orig_ilp_length}).



\section{Our Approach}


Because of the lexical diversity in student responses, we suspect the co-occurrence matrix $A$ may not establish a faithful correspondence between sentences and concepts.
A concept may be conveyed using multiple bigram expressions; however, the current co-occurrence matrix only captures a binary relationship between sentences and bigrams.
For example, we ought to give partial credit to ``bicycle parts'' (S7) given that a similar expression ``bike elements'' (S4) appears in the sentence.
Domain-specific synonyms may be captured as well. 
For example, the sentence ``I tried to follow along but I couldn't \textit{grasp the} concepts'' is expected to partially contain the concept ``understand the'', although the latter did not appear in the sentence.

The existing matrix $A$ is highly sparse. Only 2.7\% of the entries are non-zero in our dataset (\S\ref{sec:data}).
We therefore propose to \emph{impute} the co-occurrence matrix by filling in missing values.
This is accomplished by approximating the original co-occurrence matrix using a low-rank matrix. 
The low-rankness encourages similar concepts to be shared across sentences. 
The data imputation process makes two notable changes to the existing ILP framework. First, it extends the domain of $A_{ij}$ from binary to a continuous scale $[0,1]$ (Eq.~\ref{eqn:orig_ilp_bigram}), which offers a better sentence-level semantic representation. 
The binary concept variables ($z_i$) are also relaxed to continuous domain $[0,1]$ (Eq.~\ref{eqn:orig_yz}), which allows the concepts to be ``partially'' included in the summary.
Concretely, given the co-occurrence matrix $A \in \mathbb{R}^{N \times M}$, we aim to find a low-rank matrix $B \in \mathbb{R}^{N \times M}$ whose values are close to $A$ at the observed positions. Our objective function is
\begin{align}
\label{eqn:mc_obj}
\displaystyle \min_{B \in \mathbb{R}^{N \times M}} \frac{1}{2}\sum_{(i,j) \in \Omega} (A_{ij} - B_{ij})^2 + \lambda \left\| B \right\|_{*} ,
\end{align}
 
\noindent where $\Omega$ represents the set of observed value positions. $\|B\|_*$ denotes the trace norm of $B$, i.e., $\|B\|_* = \sum_{i=1}^r \sigma_i$, where $r$ is the rank of $B$ and $\sigma_i$ are the singular values. 
By defining the following projection operator $P_{\Omega}$, 
\begin{align}
\label{eqn:mc_proj_op}
\displaystyle 
[P_{\Omega}(B)]_{ij} = \left\{
     \begin{array}{lr}
       B_{ij} & (i,j) \in \Omega\\
       0 & (i,j) \notin \Omega
     \end{array}
   \right.
\end{align}

\noindent our objective function~(Eq.~\ref{eqn:mc_obj}) can  be succinctly represented as 
\begin{align}
\label{eqn:new_obj}
\min_{B \in \mathbb{R}^{N \times M}} \frac{1}{2} \| P_{\Omega}(A) - P_{\Omega}(B) \|_{F}^2 + \lambda \| B \|_{*} ,
\end{align}

\noindent where $\| \cdot \|_F$ denotes the Frobenius norm.

Following \cite{Mazumder:2010}, we optimize Eq. \ref{eqn:new_obj} using the proximal gradient descent algorithm. The update rule is 
\begin{small}
\begin{equation}
    B^{(k+1)} = \mbox{prox}_{\lambda \rho_k} \Big( B^{(k)} + \rho_k \big( P_{\Omega}(A) - P_{\Omega}(B) \big) \Big) ,
\end{equation}
\end{small}

\noindent where $\rho_k$ is the step size at iteration \emph{k} and the proximal function $\mbox{prox}_{t}(B)$ is defined as the singular value soft-thresholding operator, $\mbox{prox}_{t}(B) = U \cdot \mbox{diag}((\sigma_i - t)_+) \cdot V^\top$, where $B = U \mbox{diag}(\sigma_1,\cdots,\sigma_r)V^\top$ is the singular value decomposition (SVD) of $B$ and $(x)_+ = \max(x, 0)$.

Since the gradient of $\frac{1}{2} \| P_{\Omega}(A) - P_{\Omega}(B) \|_{F}^2$ is Lipschitz continuous with $L = 1$ ($L$ is the Lipschitz continuous constant), we follow \cite{Mazumder:2010} to choose fixed step size $\rho_k = 1$, which has a provable convergence rate of $O(1/k)$, where $k$ is the number of iterations.

\section{Dataset}
\label{sec:data}

Our dataset is collected from an introductory materials science and engineering class taught in a major U.S. university. 
The class has 25 lectures and enrolled 53 undergrad students.
The students are asked to provide feedback after each lecture based on three prompts:
1) ``{describe what you found most interesting in today's class},'' 
2) ``{describe what was confusing or needed more detail},'' 
and 3) ``{describe what you learned about how you learn}.''
These open-ended prompts are carefully designed to encourage students to self-reflect, allowing them to ``recapture experience, think about it and evaluate it"~\cite{boud:2013}.
The average response length is 10$\pm$8.3 words.
If we concatenate all the responses to each lecture and prompt into a ``pseudo-document'', the document contains 378 words on average.

The reference summaries are created by a teaching assistant.
She is allowed to create abstract summaries using her own words in addition to selecting phrases directly from the responses. 
Because summary annotation is costly and recruiting annotators with proper background is nontrivial, 12 out of the 25 lectures are  annotated with reference summaries. 
There is one gold-standard summary per lecture and question prompt, yielding 36 document-summary pairs\footnote{This data set is publicly available at \url{http://www.coursemirror.com/download/dataset}.}.
On average, a reference summary contains 30 words, corresponding to 7.9\% of the total words in student responses.
43.5\% of the bigrams in human summaries appear in the responses.

\section{Experiments}
\label{sec:experiments}

Our proposed approach is compared against a range of baselines.
They are 1) \textsc{Mead}~\cite{Radev:2004}, a centroid-based summarization system that scores sentences based on length, centroid, and position;
2) \textsc{LexRank}~\cite{Erkan:2004}, a graph-based summarization approach based on eigenvector centrality;
3) \textsc{SumBasic}~\cite{Vanderwende:2007}, an approach that assumes words occurring frequently in a document cluster have a higher chance of being included in the summary;
4) \textsc{Baseline-ilp}~\cite{Berg-Kirkpatrick:2011}, a baseline ILP framework without data imputation.

For the ILP based approaches, we use bigrams as concepts (bigrams consisting of only stopwords are removed\footnote{Bigrams with one stopword are not removed because 1) they are informative (``a bike", ``the activity", ``how materials'); 2) such bigrams appear in multiple sentences and are thus helpful for matrix imputation.}) and sentence frequency as concept weights.
We use all the sentences in 25 lectures to construct the concept-sentence co-occurrence matrix and perform data imputation. 
It allows us to leverage the co-occurrence statistics both within and across lectures. 
For the soft-impute algorithm, we perform grid search (on a scale of [0, 5] with stepsize 0.5) to tune the hyper-parameter $\lambda$.
To make the most use of annotated lectures, we split them into three folds.
In each one, we tune $\lambda$ on two folds and test it on the other fold.
Finally, we report the averaged results.
In all experiments, summary length is set to be 30 words or less, corresponding to the average number of words in human summaries.

\begin{table*}[t]
\setlength{\tabcolsep}{4pt}
\renewcommand{\arraystretch}{1.1}
\begin{small}
\begin{center}
\begin{tabular}{| l | ccc | ccc | ccc | c |}
\hline
 & \multicolumn{3}{c|}{\textbf{ROUGE-1}} & \multicolumn{3}{c|}{\textbf{ROUGE-2}} & \multicolumn{3}{c|}{\textbf{ROUGE-SU4}} & \textbf{Human}\\ 
\multicolumn{1}{|l|}{\textbf{System}} & R (\%) & P (\%) & F (\%) & R (\%) & P (\%) & F (\%) & R (\%) & P (\%) & F (\%) & \textbf{Preference}\\
\hline
\hline
\textsc{Mead} & {\cellcolor[gray]{.8}} 26.4 & {\cellcolor[gray]{.8}} 23.3 & {\cellcolor[gray]{.8}} 21.8 & {\cellcolor[gray]{.8}} \ 6.7& {\cellcolor[gray]{.8}} \ 7.6 & {\cellcolor[gray]{.8}} \ 6.3 & {\cellcolor[gray]{.8}} \ 8.8 & {\cellcolor[gray]{.8}} \ 8.0 & {\cellcolor[gray]{.8}} \ 5.4 & {\cellcolor[gray]{.8}} 24.8\%\\
\textsc{LexRank} & {\cellcolor[gray]{.8}} 30.0 & {\cellcolor[gray]{.8}} 27.6 & {\cellcolor[gray]{.8}} 25.7 & {\cellcolor[gray]{.8}} \ 8.1 & {\cellcolor[gray]{.8}} \ 8.3 & {\cellcolor[gray]{.8}} \ 7.6 & {\cellcolor[gray]{.8}} \ 9.6 & {\cellcolor[gray]{.8}} \ 9.6 & {\cellcolor[gray]{.8}} \ 6.6 & ---\\
\textsc{SumBasic} & 36.6 & {\cellcolor[gray]{.8}} 31.4 & {\cellcolor[gray]{.8}} 30.4 & \ 8.2 & {\cellcolor[gray]{.8}} \ 8.1 & \ 7.5 & {\cellcolor[gray]{.8}} 13.9 & {\cellcolor[gray]{.8}} 11.0 & {\cellcolor[gray]{.8}} \ 8.7 & ---\\
\textsc{Ilp Baseline} & 35.5 & {\cellcolor[gray]{.8}} 31.8 & {\cellcolor[gray]{.8}} 29.8 & 11.1 & {\cellcolor[gray]{.8}} 10.7 & \ 9.9 & {\cellcolor[gray]{.8}} 12.9 & {\cellcolor[gray]{.8}} 11.5 & {\cellcolor[gray]{.8}} \ 8.2 & {\cellcolor[gray]{.8}} 69.6\%\\
\textsc{Our Approach} & \textbf{38.0} & \textbf{34.6} & \textbf{32.2} & \textbf{12.7} & \textbf{12.9} & \textbf{11.4} & \textbf{15.5} & \textbf{14.4} & \textbf{10.1} & \ \textbf{89.6\%}\\
\hline
\end{tabular}
\end{center}
\end{small}
\caption{Summarization results evaluated by ROUGE (\%) and human judges. 
Shaded area indicates that the performance difference with \textsc{Our Approach} is statistically significant ($p < 0.05$) using a two-tailed paired t-test on the 36 document-summary pairs.
}
\label{tab:results}
\vspace{0.05in}
\end{table*}



In Table~\ref{tab:results}, we present summarization results evaluated by ROUGE~\cite{Lin:2004} and human judges.
ROUGE is a standard evaluation metric that compares system and reference summaries based on n-gram overlaps.
Our proposed approach outperforms all the baselines based on three standard ROUGE metrics.\footnote{F-scores are slightly lower than P/R because of the averaging effect and can be illustrated in one example. Suppose we have P1=0.1, R1=0.4, F1=0.16 and P2=0.4, R2=0.1, F2=0.16. Then the macro-averaged P/R/F-scores are: P=0.25, R=0.25, F=0.16. In this case, the F-score is lower than both P and R.}
When examining the imputed sentence-concept co-occurrence matrix, we notice some interesting examples that indicate the effectiveness of the proposed approach, shown in Table~\ref{tab:example_mc}. 

Because ROUGE cannot thoroughly capture the semantic similarity between system and reference summaries, we further perform human evaluation.
For each lecture and prompt, we present the prompt, a pair of system outputs in a random order, and the human summary to five Amazon turkers.
The turkers are asked to indicate their preference for system A or B based on the semantic resemblance to the human summary on a 5-Likert scale (`Strongly preferred A', `Slightly preferred A', `No preference',  `Slightly preferred B', `Strongly preferred B').
They are rewarded \$0.08 per task. We use two strategies to control the quality of the human evaluation. First, we require the turkers to have a Human Intelligence Task (HIT) approval rate of 90\% or above. Second, we insert some quality checkpoints by asking the turkers to compare two summaries of same text content but different sentence orders. Turkers who did not pass these tests are filtered out.
Due to budget constraints, we conduct pairwise comparisons for three systems. The total number of comparisons is  3 system-system pairs $\times$ 12 lectures $\times$ 3 prompts $\times$ 5 turkers = 540 total pairs. We calculate the percentage of ``wins'' (strong or slight preference) for each system among all comparisons with its counterparts.
Results are reported in the last column of Table~\ref{tab:results}.
\textsc{Our Approach} is preferred significantly more often than the other two systems\footnote{For the significance test, we convert a preference to a score ranging from -2 to 2 (`2' means `Strongly preferred' to a system and `-2' means `Strongly preferred' to the counterpart system), and use a two-tailed paired t-test with $p < 0.05$ to compare the scores.}. 
Regarding the inter-annotator agreement, we find 74.3\% of the individual judgements agree with the majority votes when using a 3-point Likert scale  (`preferred A', `no preference', `preferred B').

\begin{table}[t]
\setlength{\tabcolsep}{5pt}
\renewcommand{\arraystretch}{1.1}
\begin{footnotesize}
\begin{center}
\begin{tabular}{| l | r | }
\hline
\textbf{Sentence} & \textbf{Assoc. Bigrams}\\ 
\hline
\hline
\rule{0pt}{2ex}\textit{the printing} needs to better so it can & \multirow{2}{*}{\textit{the graph}}\\
be easier to read & \\\hline
\rule{0pt}{2ex}graphs make it \textit{easier to} understand  & \multirow{2}{*}{\textit{hard to}}\\
  concepts & \\\hline
\rule{0pt}{2ex}the naming system for the 2 \textit{phase} & \multirow{2}{*}{\textit{phase diagram}}\\
    \textit{regions} & \\
\hline
\rule{0pt}{2ex}I tried to follow along but I couldn't & \multirow{2}{*}{\textit{understand the}}\\
\textit{grasp the} concepts & \\
\hline
\rule{0pt}{2ex}no problems except for the specific & \multirow{3}{*}{\textit{strain curves}}\\
\rule{0pt}{2ex}equations used to determine properties& \\
\rule[-2ex]{0pt}{0pt}from the stress - \textit{strain graph} & \\
\hline
\end{tabular}
\end{center}
\end{footnotesize}
\caption{Associated bigrams do not appear in the sentence, but after Matrix Imputation, they yield a decent correlation (cell value greater than 0.9) with the corresponding sentence.
}
\label{tab:example_mc}
\end{table}

Table~\ref{tab:output} presents example system outputs. This
offers intuitive understanding to our proposed approach.

\begin{table}[th]
\setlength{\tabcolsep}{7pt}
\renewcommand{\arraystretch}{1.0}
\begin{footnotesize}
\begin{tabular}{ l }
\hline

\rule{0pt}{3ex}\textbf{Prompt}\\
\textit{Describe what you found most interesting in today's class}\\

\rule{0pt}{3ex}{\bf Reference Summary}\\
- unit cell direction drawing and indexing\\
- real world examples\\
- importance of cell direction on materials properties\\

\rule{0pt}{3ex}{\bf System Summary (\textsc{Ilp Baseline})}\\
- drawing and indexing unit cell direction\\
- it was interesting to understand how to find apf and \\
\, fd from last weeks class\\
\rule[-2.2ex]{0pt}{0pt}- south pole explorers died due to properties of tin\\

\rule{0pt}{3ex}{\bf System Summary (\textsc{Our Approach})}\\
- crystal structure directions\\
- surprisingly i found nothing interesting today .\\
- unit cell indexing\\
- vectors in unit cells\\
- unit cell drawing and indexing\\
\rule[-2.2ex]{0pt}{0pt}- the importance of cell direction on material properties\\

\hline
\end{tabular}
\caption{Example reference and system summaries. }
\label{tab:output}
\end{footnotesize}
\end{table}

\section{Related Work}
\label{sec:related_work}

Our previous work~\cite{luo-litman:2015:EMNLP} proposes to summarize student responses by extracting phrases rather than sentences in order to meet the need of aggregating and displaying student responses in a mobile application~\cite{Luo:2015:demo,fan:2015:coursemirror}. It adopts a clustering paradigm to address the lexical variety issue. In this work, we leverage matrix imputation to solve this problem and summarize student response at a sentence level.

The integer linear programming framework has demonstrated substantial success on summarizing news documents~\cite{Gillick:2008,Gillick:2009,Woodsend:2012,Li:2013:ACL}.
Previous studies try to improve this line of work by generating better estimates of concept weights. 
Galanis et al. ~\shortcite{Galanis:2012} proposed a support vector regression model to estimate bigram frequency in the summary.
Berg-Kirkpatrick et al.~\shortcite{Berg-Kirkpatrick:2011} explored a supervised approach to learn parameters using a cost-augmentative SVM.
Different from the above approaches, we focus on the co-occurrence matrix instead of concept weights, which is another important component of the ILP framework.

Most summarization work focuses on summarizing news documents, as driven by the DUC/TAC conferences.
Notable systems include maximal marginal relevance~\cite{Carbonell:1998}, submodular functions~\cite{Lin:2010:NAACL}, jointly extract and compress sentences~\cite{Zajic:2007}, optimize content selection and surface realization~\cite{Woodsend:2012}, minimize reconstruction error~\cite{He:2012}, and dual decomposition~\cite{Almeida:2013}. Albeit the encouraging performance of our proposed approach on summarizing student responses, when applied to the DUC 2004 dataset~\cite{HONG14.1093.L14-1070} and evaluated using ROUGE we observe only comparable or marginal improvement over the ILP baseline. However, this is not surprising since the lexical variety is low (20\% of bigrams appear more than twice compared to 3\% of bigrams appear more than twice in student responses) and thus less data sparsity, so the DUC data cannot benefit much from imputation.

\section{Conclusion}
\label{sec:conclusion}

We make the first effort to summarize student feedback using an integer linear programming framework with data imputation.
Our approach allows sentences to share co-occurrence statistics and alleviates sparsity issue.
Our experiments show that the proposed approach performs competitively against a range of baselines and shows promise for future automation of student feedback analysis.

In the future, we may take advantage of the high quality student responses~\cite{luo:2016:FLAIRS} and explore helpfulness-guided summarization~\cite{xiong-litman:2014:Coling} to improve the summarization performance. We will also investigate whether the proposed approach benefits other informal text such as product reviews, social media discussions or spontaneous speech conversations, in which we expect the same sparsity issue occurs and the language expression is diverse.

\section*{Acknowledgments}
This research is supported by an internal grant from the Learning Research and Development Center at the University of Pittsburgh. We thank Muhsin Menekse for providing the data set. We thank Jingtao Wang, Fan Zhang, Huy Nguyen and Zahra Rahimi for valuable suggestions about the proposed summarization algorithm. We also thank anonymous reviewers for insightful comments and suggestions.

\bibliography{references,ref_self}
\bibliographystyle{naaclhlt2016}

\end{document}